\documentclass[letterpaper, 10 pt, conference]{ieeeconf}  

\IEEEoverridecommandlockouts                              

\usepackage{graphics} 
\usepackage{epsfig} 
\usepackage{amsmath} 
\usepackage{amssymb}  
\usepackage{tensor}
\usepackage{siunitx}

\usepackage{cite}
\usepackage{url}
\usepackage{siunitx}

\usepackage{booktabs}
\usepackage{xcolor}
\usepackage[linesnumbered,ruled,vlined]{algorithm2e}
\usepackage{multirow}

\newcommand{\hide}[1]{}
\newcommand{\bdmath}{\begin{dmath}}
\newcommand{\edmath}{\end{dmath}}
\newcommand{\beq}{\begin{equation}}
\newcommand{\eeq}{\end{equation}}
\newcommand{\bdm}{\begin{displaymath}}
\newcommand{\edm}{\end{displaymath}}
\newcommand{\bea}{\begin{eqnarray}}
\newcommand{\eea}{\end{eqnarray}}
\newcommand{\beal}{\beq \begin{array}{ll}}
\newcommand{\eeal}{\end{array} \eeq}
\newcommand{\beas}{\begin{eqnarray*}}
\newcommand{\eeas}{\end{eqnarray*}}
\newcommand{\ea}{\end{array}}
\newcommand{\bit}{\begin{itemize}}
\newcommand{\eit}{\end{itemize}}
\newcommand{\ben}{\begin{enumerate}}
\newcommand{\een}{\end{enumerate}}

\title{\LARGE \bf
FAST-LIO, Then Bayesian ICP, Then GTSFM
}

\author{
Jerred Chen$^{2}$,
Xiangcheng Hu$^{1}$, 
Shicong Ma$^{2}$,
Jianhao Jiao$^{1}$,
Ming Liu$^{1}$,
and Frank Dellaert$^{2}$
\thanks{$^{1}$ Robotics Institute,  
  Department of Electronic and Computer Engineering,
  The Hong Kong University of Science and Technology \texttt{\{xhubd, jjiao\}@connect.ust.hk}, \texttt{\{eelium\}@ust.hk}}%
\thanks{$^{2}$ Institute for Robotics and Intelligent Machines,
  College of Computing, Georgia Institute of Technology
   \texttt{\{jerred, dellaert\}@cc.gatech.edu}}%
}

\begin{document}

\maketitle
\thispagestyle{empty}
\pagestyle{empty}

\begin{abstract}
For the Hilti Challenge 2022, we created two systems, one building upon the other. The first system is FL2BIPS which utilizes the iEKF algorithm FAST-LIO2 and Bayesian ICP PoseSLAM, whereas the second system is GTSFM, a structure from motion pipeline with factor graph backend optimization powered by GTSAM.
\end{abstract}

\section{OVERVIEW}

We did two separate submissions and this report describes both of them:
\begin{enumerate}
\item \textbf{FL2BIPS}: FAST-LIO2 + Bayesian ICP + Pose SLAM, based on LIDAR and IMU only.
\item \textbf{FL2BIPS-GTSFM} The same system with additionally using bundle adjustment, provided by GTSFM, a GTSAM-backed Structure from Motion pipeline.
\end{enumerate}

We start with FAST-LIO2~\cite{xu2022fast}, a filtering-based approach, followed by a batch-optimization phase in which we re-estimate the pose constraints and loop closures discovered by FAST-LIO2, using Bayesian ICP \cite{Maken2020EstimatingMU}.
The combination of FAST-LIO2 and Bayesian ICP uses only LIDAR and IMU, and is the system we used for our first set of submissions. The second set of submissions additionally runs rig bundle-adjustment on top of the PoseSLAM graph, using the GTSFM system.

We discuss both systems below.


\section{FL2BIPS}


\subsection{FAST-LIO2 (FL2)}

We employ the tightly-coupled iterated Kalman filter-based LiDAR-inertial state estimator: FAST-LIO2 \cite{xu2022fast} to estimate initial poses of sensors (in the IMU frame). 
Under the filter-based framework: motion propagation and measurement update, FAST-LIO2's pipeline can be summarized as:

\begin{enumerate}
    \item Forward propagation on motion upon each IMU measurement at a high rate (i.e., $400$ Hz).    
    \item LiDAR point de-skew: FAST-LIO2 designs the backward propagation to estimate the LiDAR pose of each point in the scan with respect to the pose at the scan end time based on IMU measurements. With these poses, each LiDAR point is transformed into the frame at the end scanning time to correct the in-scan motion.
    \item Iterated Kalman filter update: In each iteration, the de-skewed LiDAR scan is matched with the global LiDAR map by finding a set of point-to-plane correspondences. Based on the current updated state, FAST-LIO2 match the current frame points with the map points, then sensors' motion are iteratively refined by minimizing the point-to-plane residuals. 
    If the state converges, FAST-LIO2 takes the state of the last iteration as the posterior estimate.
    \item With the estimated motion, the current LiDAR scan is transformed into the global coordinate system and incorporated with the global map. FAST-LIO2 also proposes the iKD-Tree data structure, enabling incremental point insertion and deletion as well as dynamic rebalancing.
\end{enumerate}

FAST-LIO2 is an online SLAM system. The iKD-Tree design guarantees that the computation time of FAST-LIO2 is not affected by the environmental scale. On our desktop computer: Intel i7-12700K, 20-thread CPU and 64GB RAM, FAST-LIO2 takes around $15$ to \SI{30}{\ms} to process each LiDAR scan while simultaneously maintaining a global map.
The noise setting of the IMU is essential to the LIO system. 
Using an open-source tool\footnote{\url{https://github.com/ori-drs/allan_variance_ros}}, we calibrated the Allan variance given the IMU calibration sequence provided by the organizer. But the calibrated parameters did not boost the performance of the LIO system. 
Thus, we directly use the IMU noise and bias parameters from the calibration file.
From our experiments, FAST-LIO2 performs well on all sequences except for \textit{Exp03}, \textit{Exp09}, and \textit{Exp15}.
These three sequences present challenging scenarios for LiDAR-based systems: stair and narrow corridor \cite{jiao2021robust}.
This motivates for utilizing the batch estimation phase to further improve the pose estimates.

\subsection{Bayesian ICP PoseSLAM (BIPS)}

The batch estimation phase constructs a pose graph with the Bayesian ICP pose constraints and perform a batch optimization to reduce the drift error from the FAST-LIO2 pose estimation. We utilize the Bayesian ICP algorithm to generate pose constraints because Bayesian ICP can calculate the relative transform between point clouds as well as estimate the covariance online. To do this, Bayesian ICP samples the posterior distribution of the relative transform using stochastic gradient Langevin dynamics (SGLD) \cite{welling11icml_SGLD}. For this phase, the pipeline contains two steps: pose constraint generation and pose graph optimization. The following consist of major facets of the pipeline.
\begin{enumerate}
    \item Utilize a point-to-plane loss function for Bayesian ICP. For each possible edge, the module reads in the FAST-LIO2 poses output and use the FAST-LIO2 poses output as the initial estimation to perform the Bayesian ICP. 
    \item Finding the potential edges. Before we perform Bayesian ICP, we first find all the possible edges, This are calculated using the KNN for each poses. For each pose, we find the nearest 25 poses and filter the poses index pairs with step range of [5,25]. 
    \item Bad constraints filtering. We also included a constraint filter to only preserve the Bayesian ICP constraints that are similar to the FAST-LIO2 relative transform. This filter removes the Bayesian ICP constraints with large transformation error. The transformation error is calculated by comparing the translation and rotation norm of the Bayesian ICP relative transform with the FAST-LIO2 relative transform for each constraint. The threshold we are using is 4 cm for translation and 5 degrees for rotation.
\end{enumerate}

\subsection{Preintegrated IMU Factors}

In addition to the Bayesian ICP factors, we also incorporate preintegrated IMU factors as described by Forster et al.~\cite{Forster2015}. To model the bias, we detect the period in which the sensor is stationary, and introduce a new bias variable at the start of each stationary period. Hence, in effect we are modeling the bias as a piece-wise constant over the entire experiment duration. The stationary periods provide excellent "in-line" calibration that makes the IMU a more trustworthy sensor for the moving trajectories in-between the stationary periods.

\subsection{Answer to Challenge Questions}
\begin{itemize}
\item Filter or optimization-based: optimization-based, but jump-started by FAST-LIO2, which is filter-based.
\item Is the method causal? NO: while FAST-LIO2 is causal, the Pose SLAM component uses all available information in batch.
\item Is bundle adjustment (BA) used? NO: Pose SLAM is based solely on Bayesian ICP and no 3D features are optimized for.
\item Is loop closing used? YES: Fast-LIO2 gives us many proposed loop closure edges that are used both as pose constraints.
\item Exact sensor modalities used: IMU, LiDAR.
\item Total processing time for each sequence and the used hardware: Around $15$ to \SI{30}{\ms} for each LiDAR cloud in FAST-LIO2, and about \SI{1.5}{\s} to register two LiDAR clouds to generate one constraint. Multiprocessing was implemented so it took approximately 30 minutes in total to generate constraints for one sequence.
\item Whether the same set of parameters is used throughout all the sequences: For both the FAST-LIO2 and BIPS modules, we generate all results with the same set of parameters.
\end{itemize}

\section{FL2BIPS-GTSFM}

FL2BIPS-GTSFM is primarily a Bundle Adjustment (BA) pipeline jump-started by the FL2BIPS system. Below we give details on GTSFM, and then discuss how we integrate it with the FL2BIPS pipeline.

\subsection{GTSFM}

\begin{figure}
    \centering
    \includegraphics[trim=100 0 100 0, clip,width=0.90\columnwidth]{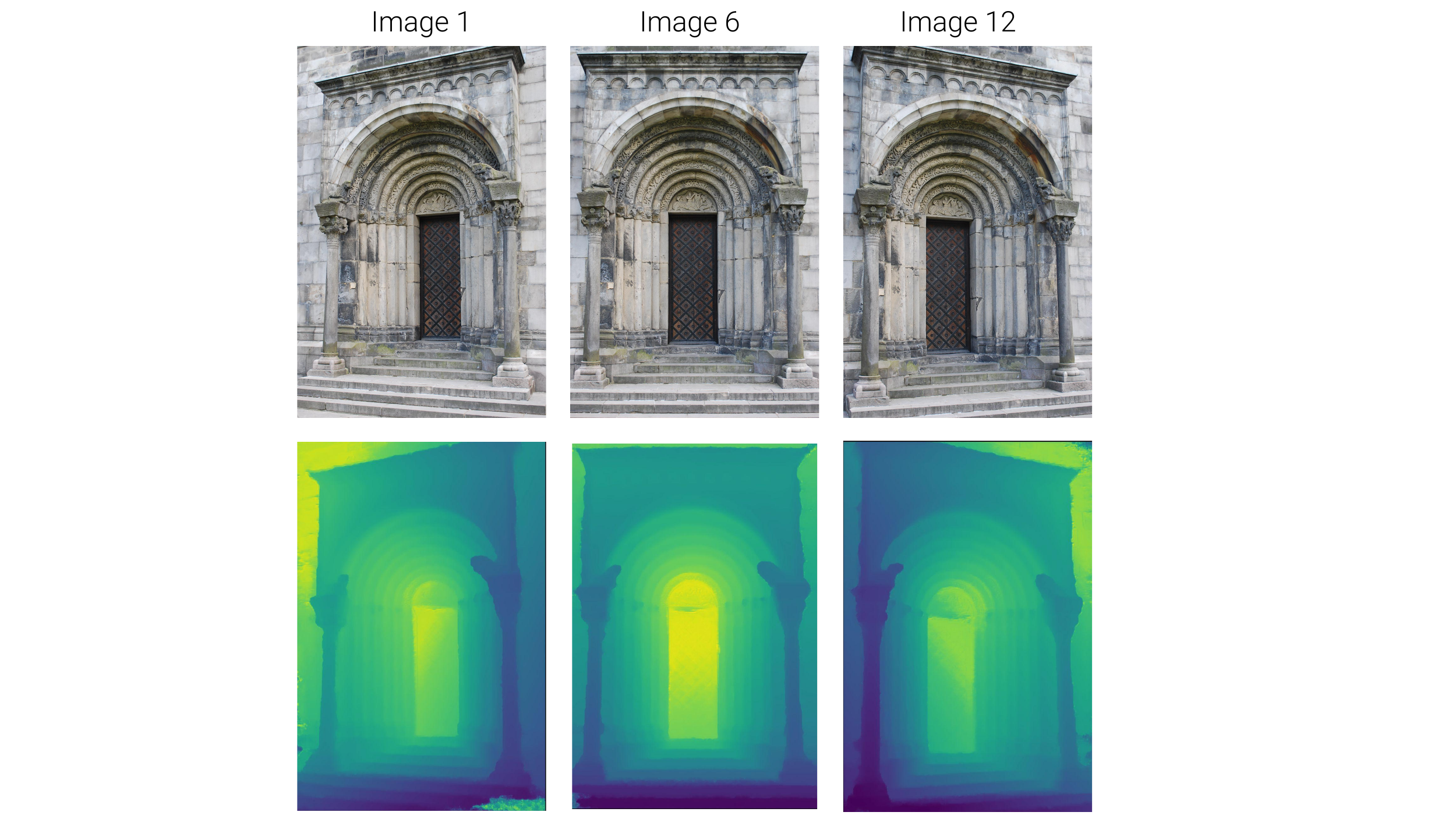}
    \includegraphics[trim=180 50 150 50, clip,width=0.90\columnwidth]{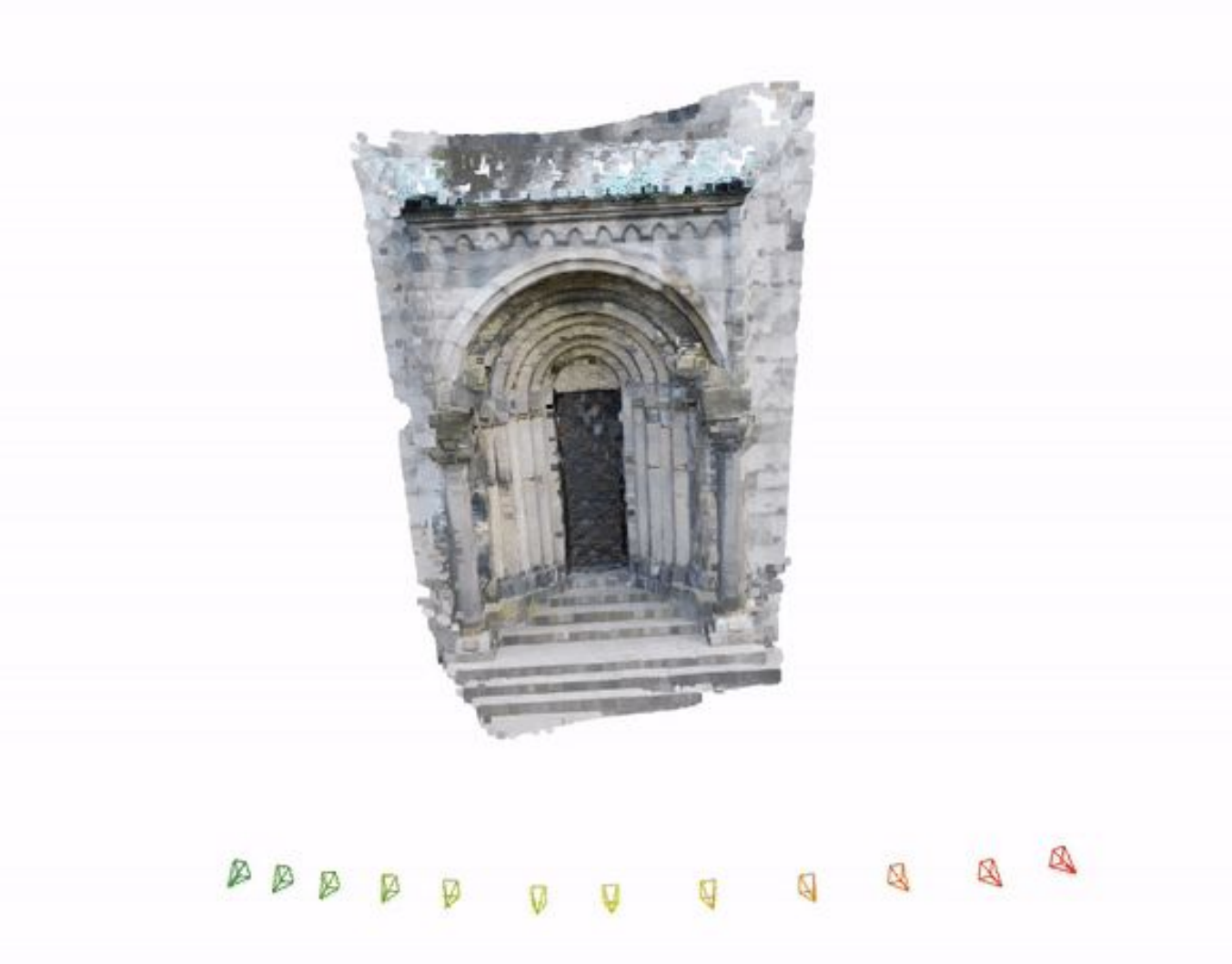}
    \caption{Qualitative results of GTSFM on the \emph{Lund Door} dataset, consisting of 12 images. \textbf{(Left)} Depth maps, generated by PatchMatchNet \cite{Wang21cvpr_PatchmatchNet}. \textbf{(Right)} Multi-view stereo output (aggregation of back-projected depth maps). }
    \label{fig:door-mvs-depth-maps}
\end{figure}

GTSFM is a distributed Structure-from-Motion pipeline that aspires to be the "COLMAP for clusters". An example of input and output is shown in Figure \ref{fig:door-mvs-depth-maps}. GTSFM uses GTSAM as the optimization back-end, and the \texttt{dask} python library to distribute computations over multiple machines. We entered this challenge precisely to push GTSFM to the limit, and have made many improvements to the system, but it is still very much a work in progress.

\begin{figure}
    \centering 
    \includegraphics[width=\columnwidth]{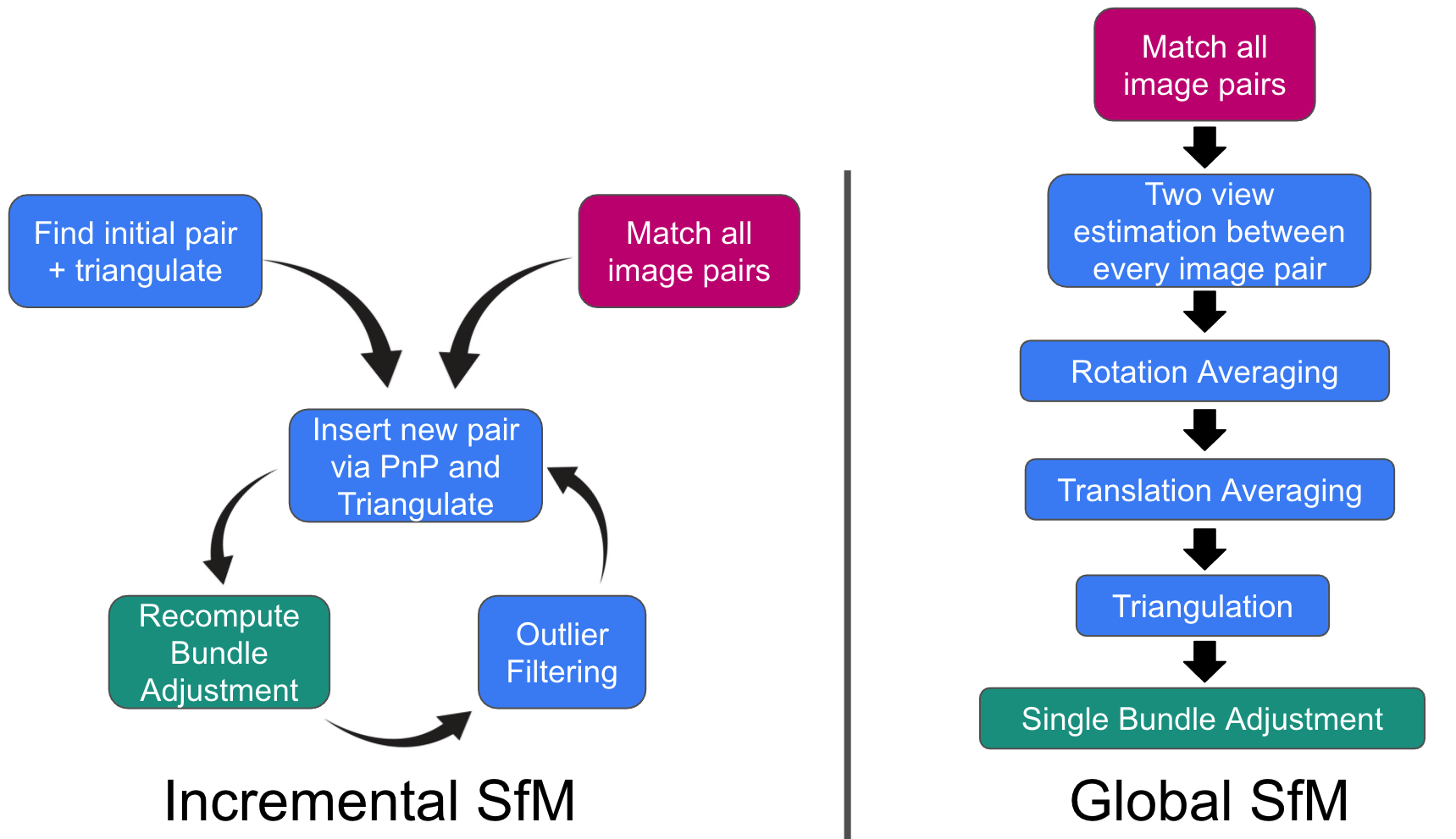}
    \caption{System diagrams of Incremental vs. Global SfM.}
    \label{fig:global}
\end{figure}

GTSFM is a new global SfM library (see Figure \ref{fig:global}) that consist of the following modules, executed in sequence but parallelized where possible over multiple workers:

\subsubsection{Front-End}

The front-end matches feature points across image-pairs that are found to have visual overlap, and robustly computes a two-view relative pose (5 dimensional due to baseline ambiguity). We use a deep feature point detector, SuperPoint \cite{Detone18cvprw_SuperPoint}, with a deep feature matcher, SuperGlue~\cite{Sarlin20cvpr_SuperGlue}, and a RANSAC verifier, which estimates a two-view . We optionally follow with two-view bundle adjustment as recommended by \cite{Julia17psivt_CriticalReviewTrifocalTensor}.

\subsubsection{Rotation Averaging}

We solve for the rotations of global camera poses using the two-view relative poses via multiple rotation averaging \cite{Hartley13ijcv_RotationAveraging}. 
We use Shonan \cite{Dellaert20eccv_Shonan}.
Shonan uses a convex relaxation based-method to guarantee convergence in the absence of outliers.

\subsubsection{Translation Averaging} 

Given camera rotations in a global frame, and pairwise translation directions,  we recover the position of each camera (translation in a global frame). We use 1dSfM \cite{Wilson14eccv_1dSfM}, which optimizes a chordal error.

\subsubsection{Data Association + Triangulation}
We use the global poses estimated by averaging to triangulate points using DLT from feature tracks, followed by a non-linear refinement.

\subsubsection{Bundle Adjustment}

\begin{figure}
    \centering
    \includegraphics[trim=50 100 0 0, clip,width=0.4\columnwidth]{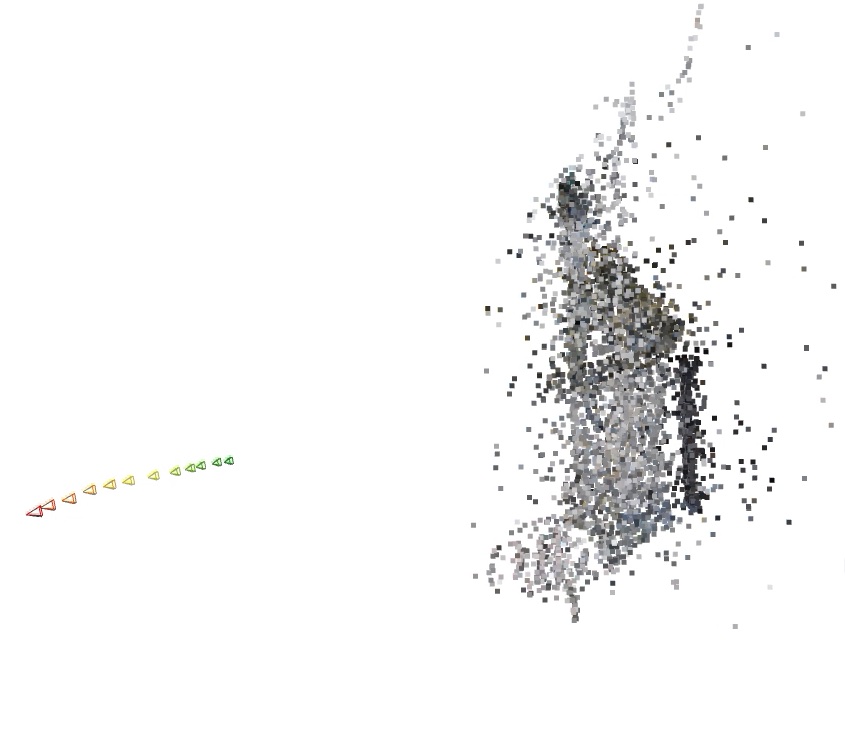}
    \hspace{10mm}
    \includegraphics[trim=0 50 0 0, clip,width=0.4\columnwidth]{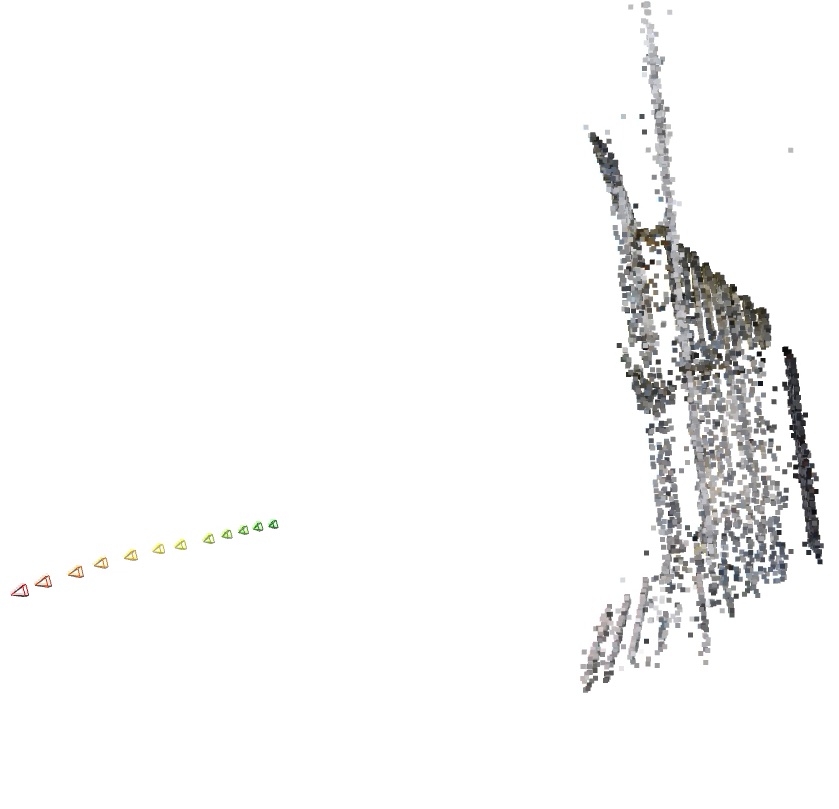}
    \caption{Qualitative results on \emph{Lund-Door-12} dataset, before \textbf{(left)} and after \textbf{(right)} bundle adjustment and filtering of 3d points by reprojection errors.}
    \label{fig:ba-input-output-door-dataset}
\end{figure}

We refine the initial camera poses and triangulated point cloud using bundle adjustment (see \ref{fig:ba-input-output-door-dataset}). We initialize the $SE(3)$ camera poses from the global camera rotations estimated from Shonan Rotation Averaging and the global camera positions estimated from 1dSfM, and refine points and camera parameters using the Levenberg-Marquardt algorithm \cite{Levenberg44qam_MethodNonlinearLeastSquares, Marquardt63siam_AlgorithmLeastSquaresNonlinear}. The back-end optimization is implemented using factor graphs~\cite{Dellaert21ar_FactorGraphs} in GTSAM~\cite{Dellaert12_GTSAM}.

\subsubsection{Multi-View Stereo}
Though not needed for the challenge, we experimented with PatchMatchNet \cite{Wang21cvpr_PatchmatchNet} and Instant NGP \cite{Mueller22arxiv_InstantNeuralGraphics}. PatchMatchNet estimates depth maps for each reference frame \emph{independently} using source images are used as evidence.  Instant-NGP \cite{Mueller22arxiv_InstantNeuralGraphics}, on the other hand, estimates 3D structure \emph{jointly} using a neural radiance field.

\subsection{Architecture Overview}

\begin{figure*}
    \centering 
    \includegraphics[width=\textwidth]{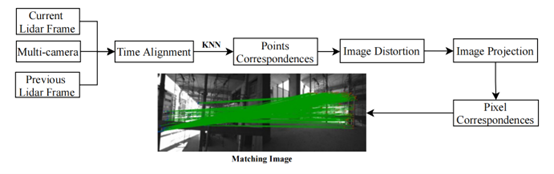}
    \caption{Proxy Correspondences Pipeline: when two scans are matched, we project the LIDAR points in all rig images associated with both poses, and count the $5^2$ overlap counts as a proxy to possible visual correspondences.}
    \label{fig:correspondences}
\end{figure*}

\subsubsection{Co-Visibility Evaluation}
The SFM framework takes initial feature pairs as input, while finding feature correspondences with less outliers across images is challenging in Hilti sequences. This is because lighting changes and texture-less appearance (such as wall, roof) are frequent in construction sites.
Thus, we propose a co-visibility evaluation module before the GTSFM part to determine a set of image pairs which share with a large co-visible region. 
The key idea is that we use LIDAR scans, which are robust to lightening and appearance changes, to evaluate the co-visibility.
The pipeline is shown in Fig. \ref{fig:correspondences}.


In particular, we use a "proxy" correspondence pipeline, which finds the number of possible matches between cameras of a pair of LIDAR scans. The pipeline of this module is shown in Figure \ref{fig: correspondences}.
The hypothesis is that for a pose pair generated from the Bayesian ICP, there should be a high possibility that the cameras associated with the pose pairs have a visual feature correspondence.
Based on this hypothesis, for each pose pair (a Bayesian ICP constraint), there will be $25$ possible camera pairs.
The key steps of this module are as follows:
\begin{enumerate}
    \item[a)] Obtaining two LIDAR point clouds is constrained by Bayesian ICP and converting these two frames to the same coordinate system. Then we perform the KNN algorithm to find nearest neighbor point pairs with a threshold of $0.05m$.
    \item[b)] Finding images that are temporally nearest to the two LIDAR frames and projecting the point pairs onto the images, respectively. Note that we only find the correspondences for the cameras across different frames.
    \item[c)] Recording the number of camera correspondences for each camera pair.
\end{enumerate}

\subsubsection{Adding Constraints in GTSFM}

GTSFM was modified to constrain all aspects of the bundle-adjustment pipeline using the constraints from the Bayesian ICP pipeline and the (assumed known) rig calibration. The ICP constraints are added as PoseSLAM constraints in the final BA factor graph and are used to constrain the rotation and translation averaging steps. The rig extrinsics are incorporated as hard constraints on the relative camera poses within a rig. We also check for visual correspondences between cameras with overlapping fields of view, implicitly performing stereo for those camera pairs.

\subsubsection{Zero-velocity Updates}
Finally, we also detect periods where the sensor is stationary by thresholding the magnitude of the constraints coming from ICP. We do not attempt any visual correspondences for those stretches and discard the ICP constraints, opting instead for hard equality constraints that prevent drift in these intervals.

\subsection{Answer to Challenge Questions}
\begin{itemize}
\item Filter or optimization-based: optimization-based, but jump-started by FAST-LIO2, which is filter-based.
\item Is the method causal? NO: while FAST-LIO2 is causal, the Pose SLAM component and GTSFM use all available information in batch.
\item Is bundle adjustment (BA) used? YES, full bundle adjustment, including global rotation and translation averaging for initialization.
\item Is loop closing used? YES: FL2BIPS gives us many proposed loop closure edges used both as pose constraints (as in FL2BIPS) and being checked for visual correspondences.
\item Exact sensor modalities used: IMU, LIDAR, and all camera images. Sparse stereo is done implicitly as we constrain the relative poses of overlapping cameras, but we do not treat it separately otherwise. No dense stereo was used.
\item Total processing time for each sequence and the used hardware: impossible to answer precisely, but hours per sequence. In addition to FL2BIPS, feature extraction and detection takes about 30 minutes per sequence, and two-view correspondence can take 2 hours. Sequences were run on different computers, mostly laptops.
\item is the same set of parameters used throughout all the sequences? YES.
\end{itemize}

\section*{ACKNOWLEDGMENTS}

Besides the core team, we are indebted to many collaborators to FASTLIO, GTSAM, and GTSFM. For GTSFM, Ayush Baid, Akshay Krishnan, Jonathan Womack, and Travis Driver. For GTSAM, Fan Jiang, Varun Agrawal, and many others. John Lambert, in particular, was instrumental in developing GTSFM and contributed text and figures to this document.


\bibliographystyle{IEEEtran}
\bibliography{refs.bib}

\end{document}